\documentclass{article}

\usepackage{microtype}
\usepackage{graphicx}
\usepackage{booktabs} 

\usepackage{hyperref}

\usepackage[accepted]{synsml2023}

\usepackage{amsmath}
\usepackage{amssymb}
\usepackage{mathtools}
\usepackage{amsthm}

\usepackage[capitalize,noabbrev]{cleveref}

\theoremstyle{plain}
\newtheorem{theorem}{Theorem}[section]

\theoremstyle{definition}

\theoremstyle{remark}

\usepackage[textsize=tiny]{todonotes}

\usepackage{caption}
\usepackage{subcaption}

\synsmltitlerunning{Group Invariant Global Pooling}

\begin{document}

\twocolumn[
\synsmltitle{Group Invariant Global Pooling}



\synsmlsetsymbol{equal}{*}

\begin{synsmlauthorlist}
\synsmlauthor{Kamil Bujel}{equal,cam}
\synsmlauthor{Yonatan Gideoni}{equal,cam}
\synsmlauthor{Chaitanya K. Joshi}{cam}
\synsmlauthor{Pietro Liò}{cam}
\end{synsmlauthorlist}

\synsmlaffiliation{cam}{University of Cambridge}

\synsmlcorrespondingauthor{Kamil Bujel}{kdb36@cl.cam.ac.uk}
\synsmlcorrespondingauthor{Yonatan Gideoni}{yg403@cl.cam.ac.uk}

\synsmlkeywords{Equivariance, Invariance, Geometric Deep Learning, Group-Equivariant Convolutional Neural Networks, Orbits}

\vskip 0.3in
]



\printAffiliationsAndNotice{\synsmlEqualContribution} 

\begin{abstract}
Much work has been devoted to devising architectures that build group-equivariant representations, while invariance is often induced using simple global pooling mechanisms. Little work has been done on creating expressive layers that are invariant to given symmetries, despite the success of permutation invariant pooling in various molecular tasks. In this work, we present Group Invariant Global Pooling (GIGP), an invariant pooling layer that is provably sufficiently expressive to represent a large class of invariant functions. We validate GIGP on rotated MNIST and QM9, showing improvements for the latter while attaining identical results for the former. By making the pooling process group orbit-aware, this invariant aggregation method leads to improved performance, while performing well-principled group aggregation.

\end{abstract}

\begin{figure*}[t!]
    \centering

     \begin{subfigure}[t]{0.49\textwidth}
         \centering
         \includegraphics[width=0.8\textwidth]{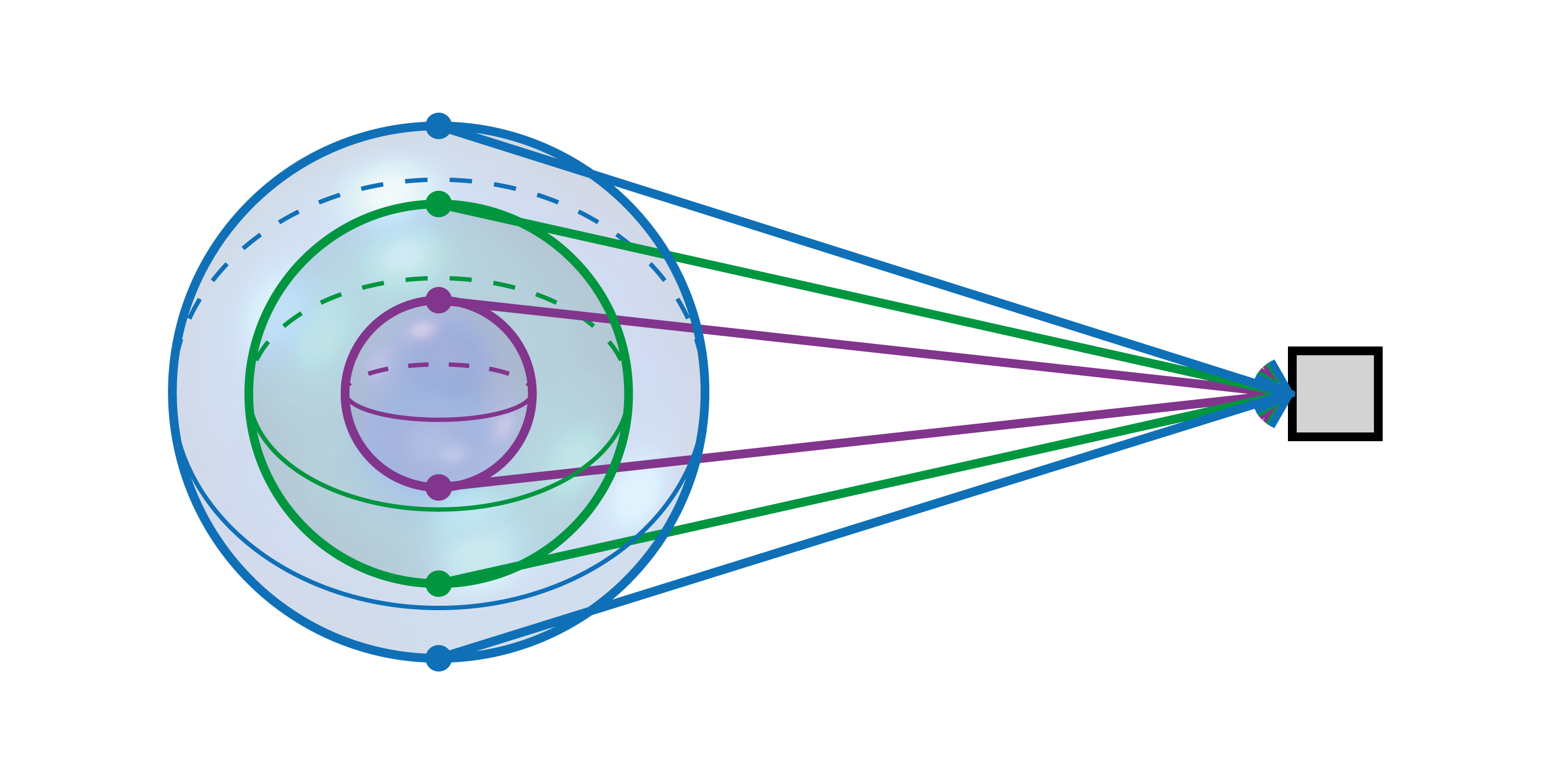}
     \end{subfigure}
     \hfill
     \begin{subfigure}[t]{0.49\textwidth}
         \centering
         \includegraphics[width=0.8\textwidth]{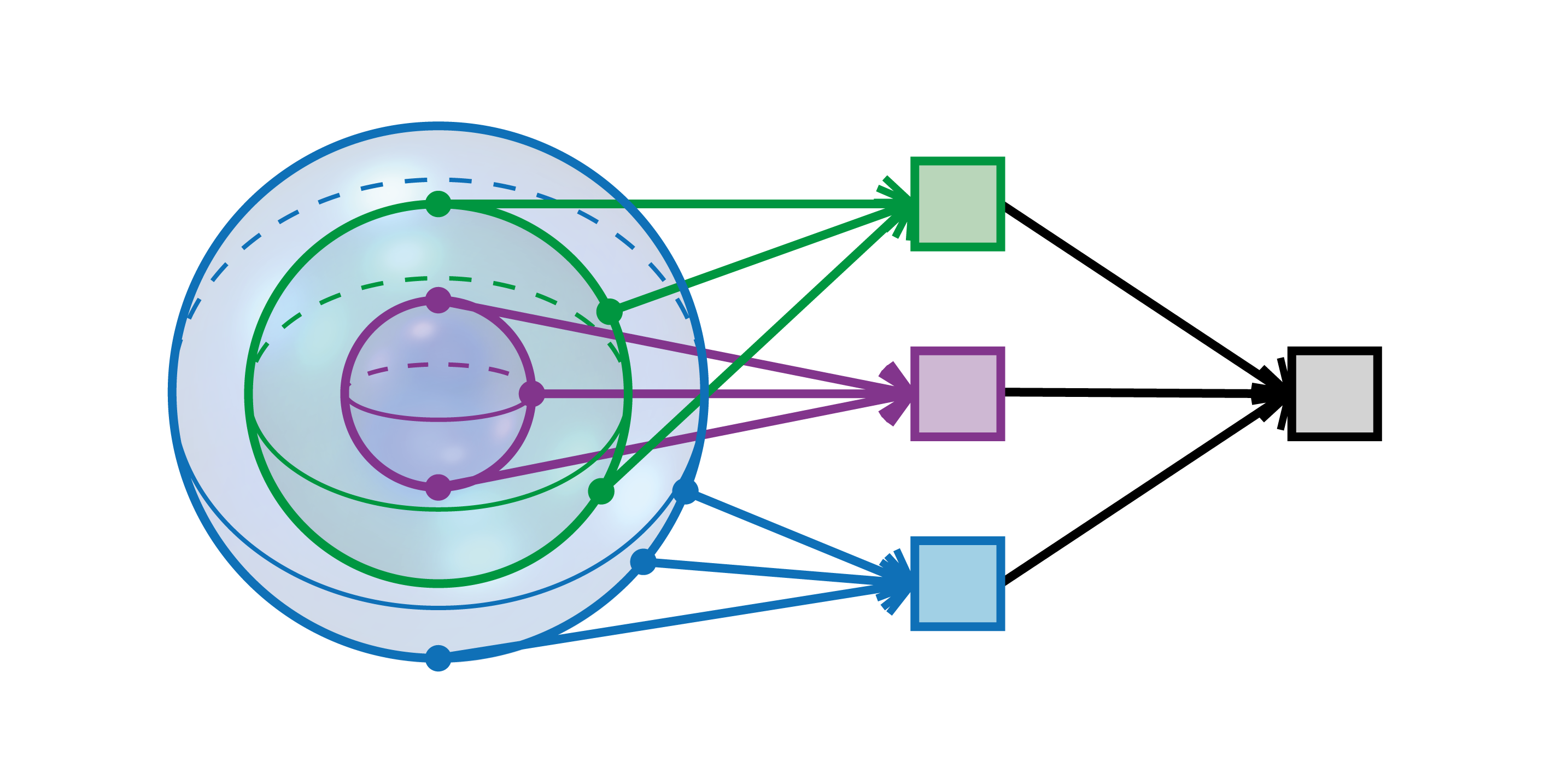}
     \end{subfigure}
     \caption{Visualization of the standard global pooling method (left) and our proposed Group-Invariant Global Pooling (right). Each sphere represents an orbit in $SO(3)$. GIGP builds group-invariant representations by aggregating each orbit separately, unlike global pooling. This leads to better-principled aggregation that is designed to respect the given symmetry.}
        \label{fig:three graphs}
\end{figure*}

\section{Introduction}
The need for a solution to have an essence of naturalness dates back to antiquity in the sciences. This line of thought led to Kepler's theory relating the radii of the planets and the Platonic solids in his \textit{Mysterium Cosmographicum} \citep{kepler_1596}, the categorization of particles in physics using their symmetries \citep{1961Neeman,gellman}, and Newton's third law \citep{newton1687philosophiae}. The notion of a change naturally resulting in an equivalent effect elsewhere found its way into deep learning as \textit{equivariance}, where transforming an input leads to an equal change in the output. 

Incorporating equivariance into architectures is a difficult task, with much of geometric deep learning being devoted to doing so \citep{bronstein2021geometric}. While architectures with specific equivariances have been very successful \citep{scarselli2008graph,sanchez2020learning, joshi2023expressive}, their construction involves highly specific computational primitives for each symmetry. Much effort has been devoted to building architectures that are equivariant to generic pre-specified symmetries \citep{cohen2016group,li2018deep,weiler2019general,finzi2020generalizing}.

Leveraging symmetries in deep learning has yielded state-of-the-art results for various tasks. Scientific problems are prime examples, as they exhibit an abundance of natural symmetries \citep{Jumper2021HighlyAP, Strk2022EquiBindGD, wang22aa}. One architecture that was used for this purpose is LieConv \citep{finzi2020generalizing}, an architecture that is equivariant to generic continuous Lie groups. \citet{finzi2020generalizing} demonstrate its efficacy on tasks such as classifying rotated images, predicting molecule's properties, and simulating physical systems.

However, most recent work has been devoted to the design of equivariant layers, while little focus was put on building expressive invariant layers. Invariance is often achieved through global or mean pooling the final layer's results. This causes loss of information about the structure of the data, as the affinity between points on the same orbits is not used. Following recent developments on invariant pooling \citep{Laptev2016TIPOOLINGTP,Yang2020BreakingTE,Azizian2020ExpressivePO} and to alleviate this loss of information, we present \textbf{Group Invariant Global Pooling} (\textbf{GIGP}), an orbit-aware invariant pooling layer, with \textbf{our contributions being as follows:}
\begin{enumerate}
    \item We propose GIGP, group invariant global pooling, and prove that any invariant function that fulfills certain conditions can be expressed as a decomposition identical to GIGP formulation.
    \item We use GIGP as a pooling method for equivariant representations from LieConv and evaluate it on rotated MNIST \citep{larochelle2007empirical} and QM9 datasets. GIGP's orbit-aware invariant pooling leads to improved results on the QM9 task, while achieving comparable performance on the rotMNIST.
\end{enumerate}

\section{Background}
\subsection{Groups and Equivariance }
Most natural systems can be described by features that remain unaltered under certain actions. 
They can be rigorously expressed using group theory. A group $(G, \circ)$ is a set $G$ of elements $u\in G$ together with a closed associative binary operation $\circ:G \times G\rightarrow G$, an inverse $u^{-1} \in G$ and the identity element $e$. A canonical example is the rotation group $SO(2)$, which denotes all two-dimensional rotations in Euclidean space. 

These abstract group elements can be represented by matrices and vectors in a way that retains their structure. For some groups each element $u \in G$ can be represented using a mapping $\rho: G \rightarrow GL(n)$ that associates it to an invertible matrix $\rho(u) \in \mathbb{R}^{n\times n}$. This representation must be a homomorphism so it retains the group's structure. This allows expressing group elements as transformations over $\mathbb{R}^n$, thus making them easier to study. For $SO(2)$ the representation $\rho(u)$ yields the well known rotation matrices $R(t) = \big(\begin{smallmatrix} \cos t & \sin t \\ -\sin t & \cos t\end{smallmatrix}\big)$ for an angle $t \in \mathbb{R}$.

In particular, group representations allows studying how functions change when transforming their inputs. An important class is \textbf{G-equivariant} functions, functions where a transformation in their input leads to a corresponding change in their output. Formally, $f: X \rightarrow Y$ is G-equivariant iff

\begin{equation}
    \forall x \in X, g \in G:\,\, f(\rho_X(g)(x)) = \rho_Y(g)f(x),
\end{equation}
where $\rho_X,\rho_Y$ denote different representations of the group element. Invariance is then a special case when $\rho_Y = I$.

\subsection{Group Convolution}



The notion of equivariance through convolutions was generalized by \citet{cohen2016group} to groups, yielding group convolutions. This allows the definition of convolutions that are equivariant to generic symmetries \citep{kondor2018generalization}, with them in this case being over functions of group elements $k,f:G\to \mathbb{R}$. It is defined as
\begin{equation}
    (k*f)(u)\equiv\int_G{k(v^{-1}u)f(v)d\mu(v)}
    \label{eq:conv}
\end{equation}
where $d\mu(v)$ is the Haar measure, a measure that allows consistently defining integrals over groups.

\subsection{Lifting}

As most group-equivariant architectures operate on Euclidean coordinates $x \in \mathbb{R}^n$, not the group elements $u \in G$, one needs to convert coordinates to group elements. To do so, a \textit{lifting} method $\text{Lift$(x)$}$: $  \{(x_i, f_i)\} \rightarrow \{(u_{ik}, f_i) \}$ is applied to convert coordinates into group elements \citep{kondor2018generalization},
\begin{equation}
    \text{Lift($x$)} = \{u \in G: uo = x \},
\end{equation}
where $o$ is a chosen origin that maps to the identity. Each coordinate $x_i$ is lifted into $K$ group elements $u_{ik}$ so it becomes non-ambiguous. For example, $K=1$ for $SO(2)$ or $K=2$ for $\mathbb{R} \times$ $SO(2)$.

However, standard lifting can lead to loss of information. For instance, given a pair of points sampled from $2$ different rotation symmetric equations,\footnote{e.g. $4 = x^2 + y^2$ and $9 = x^2 + y^2$} their lifted representations only encodes the rotation angle. \citet{finzi2020generalizing} propose to encode group \textit{orbits} during the lifting process. Intuitively, orbits represent a set of points that can all be reached through group transformations from the same origin $o$. Formally, a quotient space $Q = X / G$ contains all orbits $q \in Q$ of $G$ present in the input coordinates $X \subset \mathbb{R}^n$, which are the information the lifting discards. To fix this, the lifting procedure is performed separately for each orbit to return pairs $(u_i, q_i) \in G \times Q$ that contain the group element $u_i$ and the orbit identifier $q_i$. This ensures that no information is lost.

\subsection{LieConv}
To facilitate learning over Lie groups, \citet{finzi2020generalizing} proposed \textit{LieConv}, an architecture that uses Lie theory to perform group-equivariant convolutions over arbitrary Lie groups. This is done by mapping the input data, expressed as pairs of coordinates and values $\{(x_i, f_i)\}\, x_i \in \mathbb{R}^n$, to corresponding group elements and efficiently convolving them. A summary of this procedure is given below.



While technically group convolution could be performed as per Equation \ref{eq:conv}, with the kernel $k$ being an MLP, this would yield poor results. As canonical neural networks perform poorly on non-Euclidean data \citep{bronstein2021geometric}, one must convert the group elements to a Euclidean domain. The Lie algebra $\mathfrak{g}$ is one such space, and the group elements can be converted to it using the logarithm map $a = \log(u)$.


This lifting procedure and mapping to the Lie algebra elements results in the LieConv convolution equation
\begin{equation}
    h(u, q_u) = \int_{v \in \text{nbhd($u$)}} k(\log(v^{-1}u), q_u, q_v)f(v, q_v)d\mu(v)
    \label{eq:lieconv}
\end{equation}
where the integral is over a neighborhood of points to ensure locality, analogous to standard convolutions in deep learning. As the exact integral is intractable, it is approximated using a Monte-Carlo estimate, thus yielding only approximate equivariance. This assumes that the group elements in are locally distributed proportionally to the Haar measure.

Akin to standard convolutional networks, multiple layers of LieConv are then combined into a ResNet architecture \cite{he2016deep}, with global pooling as the last layer.

\section{Group Invariant Global Pooling}
\begin{table*}[t]
    \centering
    \caption{Results when appending GIGP to LieConv on various tasks. rotMNIST has an $SO(2)$ symmetry while pre-processed QM9 has $SO(3)$.}
    \label{tab:results}
    \begin{tabular}{lcccc}
        & \multicolumn{2}{c}{rotMNIST} & \multicolumn{2}{c}{QM9 $\epsilon_{HOMO}$} \\
        \midrule
        & Training Error & Test Error & Train MAE & Test MAE \\
        & \% & \% & meV & meV \\
        \midrule
        \multicolumn{1}{l}{LieConv + Simple Global Pooling}  & $0.05 \pm 0.05$ & $1.42 \pm 0.13$ & $4.8$ & $30.6$ \\
        \multicolumn{1}{l}{LieConv + GIGP} & $0.03 \pm 0.05$ & $1.38 \pm 0.07$ & $4.7$ & $29.4$          
    \end{tabular}
\end{table*}
While the equivariant convolutions make the model respect a symmetry, they do not assure that the model's predictions are invariant. Thus, it is natural to introduce a group invariant global pooling layer, analogous to the aggregation functions in standard graph neural networks (GNNs). To make invariant global predictions it is sufficient to build invariant representations of each orbit. In this work, we propose a general framework to generate such orbit-aware invariant features.

We decompose a sum over the group elements $v \in G$ into a sum over orbits $q \in Q$ and group elements within each orbit $u \in G^q$:
\begin{equation}
    \sum_{v \in G} f(v)=\sum_{q\in Q} \sum_{u\in G^q} f(u)
\end{equation}
where $u \in G^q \subset G$ such that $[u, q] \in G \times Q$, indicating the $u$ is observed within the orbit $q$. The inner sum is invariant to the group transformations if $f(u)$ is equivariant, as transformed elements always stay within the same orbit.

We can now use this decomposition to obtain a richer aggregation method that uses an MLP $\Phi$ to build invariant representations of orbits. We call this group invariant global pooling \textbf{(GIGP)}, defined as

\begin{equation}
    \text{GIGP}(f,G)=C\sum_{q\in Q} w_q\Phi\left(\sum_{u\in G^q} f(u), q\right)
\end{equation}

where $w_q$ are learnable parameters and $\Phi$ is a neural network. $C$ is a constant that can be tuned to make GIGP, for example, initialized as a standard pooling layer. $w_q$ are used to weigh orbits according to their relevance.

As most architectures use a global mean-pooling aggregation layer, $\Phi$ and $C$ are initialized so that GIGP before training performs global mean-pooling. This was found to improve results in practice. Moreover, to allow soft invariance \cite{van2022relaxing}, nearby orbits are aggregated together based on their distance from the main orbit $q$. This allows the application of GIGP to sparse data, such as point clouds and molecules.

Finally, we prove GIGP's expressivity by showing that any invariant continuous function has an identical decomposition to GIGP preceded by an equivariant network and followed by an MLP. To prove it, we generalize the Deep Sets proof by \citet{Zaheer2017DeepS} from permutation invariant operations to arbitrary non-homogeneous groups. The proof and exact theorem are given in Appendix \ref{app:proof}.

\begin{theorem}
\text{(Sufficient Expressivity of GIGP, Informal)} Let $f:X\to F$ be some invariant function that takes a set of group elements as input. Then, there exist functions $\rho,\phi,\psi$ such that $f(X)\equiv\rho(\sum_{q\in Q}\phi[\sum_{x\in q}\psi(x)])$.
\end{theorem}
For simplicity, we elect to use GIGP without an appended network, thereby setting $\rho$ to the identity. Showing that this form is also sufficient is left for future work.

\section{Experiments}
To evaluate GIGP, we combine it with LieConv and train on rotMNIST and QM9 datasets, with further investigation against more architectures and tasks left for future work due to computational constraints.

All experiments were run on 16GB Nvidia Tesla P100 GPUs. The rotMNIST/QM9 models took on average $4/16$ hours respectively to train for $500$ epochs. For both architectures the hyperparameters  given in \citet{finzi2020generalizing} were used, which are a learning rate of $0.003$, a batch size of $32$, and a limit on each point's neighborhood to $32$ nearest neighbors for rotMNIST and batch size of $75$ and neighborhood of $100$ for QM9. Hyperparameter tuning was not performed and is left for future work. $SO(n)$ was used as the symmetry group for both models, being the simplest non-homogeneous symmetry both exhibit. The models are implemented in PyTorch\footnote{For code please contact the authors.} \citep{paszke2017automatic}. The results are shown in Table \ref{tab:results}.

\subsection{Rotated Images}

We use the rotMNIST dataset \citep{larochelle2007empirical}, which contains rotated images of digits from the popular image recognition MNIST task \citep{deng2012mnist}. These images are invariant under the rotation group $SO(2)$. We use the original splits of $10$k for training, $2$k for validation, and $50$k for the test set.

To test GIGP, we evaluate it together with LieConv, replacing its global mean pooling layer. By running with original hyperparameters and training procedure, we achieve comparable results, as shown in Table \ref{tab:results}. We believe this result on rotMNIST, a saturated task, validates that our method is not only as expressive as standard pooling, but does not impair training.

\subsection{Molecules}
We also evaluate our proposed pooling method on the QM9 molecular property task \citep{wu2018moleculenet}. The dataset contains molecules as 3D coordinates of atoms and their respective atomic charges. The coordinates have no canonical origin, therefore they are $E(3)$ equivariant and well-performing models must use that geometric prior. To incorporate $SO(3)$ symmetry, all molecules were re-centered. We report results on the \texttt{Homo} task of predicting the energy of the highest occupied orbital. The dataset contains $100$k molecules in the train set, $10\%$ in the test set and the remaining ones in the validation set, as per \citet{Anderson2019CormorantCM}. As shown in Table \ref{tab:results}, the resulting GIGP train/test errors are slightly improved over vanilla Global Pooling. These results are promising, as they were achieved with no hyperparameter tuning.

\section{Related Work}
\paragraph{Group-Equivariant Convolution}
\citet{lecun1995convolutional} introduced convolutional neural networks, the first translation equivariant architecture.  \citet{cohen2016group} generalized this notion of equivariance to arbitrary groups.  Later works extended this by using irreducible group representations to perform convolution with respect to various continuous groups and their discrete subgroups. This includes $SO(2)$ \citep{cohen2016steerable}, $O(2)$ \citep{weiler2019general}, $SO(3)$ \citep{thomas2018tensor} and $O(3)$ \citep{geiger2022e3nn} groups.

However, there is a lack of work that attempts to generate architectures equivariant to any symmetry group. \citet{ravanbakhsh2017equivariance} achieve equivariance to any finite group by sharing weights over orbits. \citet{lang2020wigner} present steps necessary to derive convolution kernels for arbitrary compact groups. \citet{van2020mdp} propose an explicit method to compute equivariant layers in the context of reinforcement learning, but their method proves too complex for larger or continuous groups. \citet{finzi2020generalizing} present LieConv, which is equivariant to arbitrary Lie groups and scales well to larger datasets.

\paragraph{Group Invariant Pooling}
\citet{laptev2016ti} introduce transformation-invariant pooling for CNNs. \citet{cohen2016group} propose coset pooling that is equivariant to any discrete group. \citet{sosnovik2019scale} propose scale-invariant pooling in CNNs. \citet{van2022relaxing} obtain G-invariant global pooling for non-stationary kernels. \citet{xu2021group} propose group equivariant/invariant subsampling for discrete groups. However, there is no work that implements group invariant global pooling for arbitrary groups. We attempt to fill this gap.



\section{Conclusion}
We presented GIGP, an expressive pooling layer that is invariant to generic symmetries. It builds group-invariant representations of all elements on the same orbits, leading to pooling operation that is better aligned with the task at hand. This mechanism addresses the lack of invariant pooling methods for group-equivariant architectures. We proved GIGP is sufficiently expressive to model a large class of invariant functions and tested it on popular image and molecular benchmarks.


\section*{Broader impact}
This work is anticipated to improve equivariant models that handle physical and dynamical systems. We hope it allows to better express group-invariant properties of those tasks, while also being more parameter efficient. Thereby, we do not anticipate it to directly have a negative impact, and hopefully the contrary --- attaining better results on such tasks could lead to scientific discoveries that benefit humanity.

\section*{Acknowledgements}
We would like to thank Petar Veličković for helpful comments and discussions. We would also like to thank Jonas Jürß and Dulhan Jayalath for comments on an earlier version of this work.




\nocite{langley00}

\bibliography{example_paper}
\bibliographystyle{synsml2023}

\newpage
\appendix
\onecolumn
\section{Proof of GIGP Expressivity}
\label{app:proof}
\begin{theorem}
    Let $f:F^X\to\mathbb{R}$ be some function that is invariant to a group's action, so $\forall g\in G:f(gx)=f(x)$. Here $F$ is some set representing the possible values given to each group element. Thus, assuming:
    \begin{enumerate}
        \item $X, F$ are countable.
        \item There are a finite number of orbits.
    \end{enumerate}
    Given these assumptions, there exist functions $\rho,\phi,\psi$ such that
    $f(X)=\rho(\sum_{q\in Q}\phi[\sum_{x\in q}\psi(x)])$, where with some abuse of notation we denote the value given to each group element over the domain with the group element directly.
\end{theorem}
\textit{Proof:}
The proof goes as follows. We shall show that there exists $\psi$ such that $\sum_{x\in q}\psi(x)$ is group invariant and constitutes an invertible function from $F^X$ to a countable set. Afterwards we'll show that there exists a function $\phi$ such that $\sum_{q\in Q}\phi(h(q))$ is an invertible function from $Q^\mathbb{N}$ to a countable set. Finally, we'll show that we can construct a function $\rho$ such that $f(X')=\rho(\sum_{q\in Q}\phi[\sum_{x\in q}\psi(x)])$. 

Step 1: As $F$ is a countable domain, we have that $\exists c:c(x)$ is an invertible mapping to $\mathbb{N}$. Because $X$ is countable each orbit has countably many elements. Thus, one can define $L(x):=\ln(p_{c(x)})$, where $p_i$ denotes the $i$-th prime. Note that this constitutes a one-to-one correspondence between sets $\{x\}_{x\in q}$ and logarithms of numbers as each number has a unique prime representation and $\sum_{x\in q}L(x)=\ln(\prod_{x\in q}p_{c(x)})$. Thus, we have shown that there exists a function $\psi(x):=\ln(p_{c(x)})$ that is an invertible mapping between $F^X$ and a countable set, specifically $\ln(\mathbb{N})$. This is trivially group invariant as the action of the group simply permutes the set.

Step 2: We denote an arbitrary function from an orbit to the real numbers as $h:Q\to\mathbb{R}$. We want to show that there exists a function $\phi$ such that $\sum_{q\in Q}\phi(h(q))$ is invertible from $Q^\mathbb{N}$, all the multisets of the form $\{\{h(q)\}\}_{q\in Q}$ to $\mathbb{N}$. As there is a finite number of orbits, one can use a similar construction to the previous step - $h(q)$ can be treated as a natural number (for example, calculating $\exp(h(q))$ if following the previous construction) and define $\phi(n):=\ln(p_n)$.

Step 3. Let $g(X):=\sum_{q \in Q}\phi[\sum_{x\in q}\psi(x)]$. This is an invertible group-invariant function from $F^X$ to $\ln(\mathbb{N})$. Thus, setting $\rho:=f\circ g^{-1}$ completes the proof.
\qed


\end{document}